\documentclass[conference]{IEEEtran}
\IEEEoverridecommandlockouts
\usepackage{cite}
\usepackage{amsmath,amssymb,amsfonts}
\usepackage{algorithmic}
\usepackage{graphicx}
\usepackage{textcomp}
\usepackage{xcolor}
\usepackage{url}
\usepackage{flushend}

\usepackage{multirow}
\usepackage{booktabs}

\def\BibTeX{{\rm B\kern-.05em{\sc i\kern-.025em b}\kern-.08em
    T\kern-.1667em\lower.7ex\hbox{E}\kern-.125emX}}
\begin{document}

\title{Self-Representation Alignment in Diffusion Models for Video Background Music Generation\\

}

\author{
\IEEEauthorblockN{
Jincheng Zhang, Gy\"orgy Fazekas, Charalampos Saitis
}

\IEEEauthorblockA{
Centre for Digital Music, Queen Mary University of London, UK
}
}

\maketitle

\begin{abstract}
Diffusion models have been widely used for various generative tasks. However, recent studies have demonstrated that the internal representations learned in the diffusion models' denoising process are insufficiently meaningful, which can hinder the generation performance. To address this, we propose a novel diffusion U-Net architecture enhanced by explicitly aligning the weaker representation from the earlier layer conditioned on higher noise with the better representation from the later layer conditioned on lower noise during training, which we use for the challenging multimodal task of video background music generation. Our diffusion U-Net Self-Representation Alignment (U-SRA) employs an alignment loss to effectively guide our diffusion model to learn good internal representations, thereby enabling it to produce higher-quality symbolic music conditioned on video features. We employ a cross-attention mechanism to synchronize music with video segments. Experiments on a benchmark video-music dataset demonstrate that the proposed U-SRA approach for video background music generation surpasses the state-of-the-art across key evaluation metrics, including video–music alignment and overall music quality. 
\end{abstract}


\section{Introduction}
\label{sec:intro}
With the increasing prevalence of videos on social media and streaming platforms such as TikTok and YouTube, background music generation has emerged as a compelling application of multimodal generative modelling. Background music can significantly enhance the emotional impact and storytelling of a video, making the content more engaging for viewers \cite{b1}. However, non-expert digital content creators often struggle to select or create a piece of music that perfectly matches a given video. As a result, there is growing interest in automated video background music generation, where deep learning models produce music tailored to the visual content both semantically and temporally.

Music composition traditionally requires deep musical knowledge and expertise. In recent years, deep learning has demonstrated the ability to learn complex musical patterns from large music datasets to generate new compositions, leading to its widespread adoption in the field of music generation \cite{b2,b3}. Generally speaking, music generation can be categorized into symbolic music production and audio synthesis. The former refers to the automatic production of music in a symbolic format, such as the Musical Instrument Digital Interface (MIDI), whereas audio synthesis directly generates sound signals. This work explores symbolic music generation using deep learning techniques, focusing on the compositional aspects of music rather than the acoustical properties emphasized in audio synthesis. Compared to the latter, generating symbolic music has the advantages of being both interpretable and editable within standard digital audio workstations as symbolic music is represented as discrete tokens corresponding to clear musical note attributes such as pitch, duration, and velocity.

In the past few years, diffusion models have emerged as a dominant approach for diverse generative tasks with continuous data such as images and audio, but their potential for discrete domains like symbolic music remains largely under-explored. Although a few prior studies have investigated the use of diffusion models for symbolic music generation, these approaches often necessitate an auxiliary VAE to embed symbolic music into continuous embeddings \cite{b4}. Moreover, diffusion models typically supervise only the final outputs during training and struggle to effectively learn the internal representations, which is a primary bottleneck in training diffusion models for generation \cite{b5}. In this work, we propose a novel U-Net-based diffusion model for symbolic music generation conditioned on videos. It enables self-representation alignment (SRA) to provide guidance by aligning the diffusion U-Net's latent representation from the earlier
layer conditioned on higher noise with that from the later layer conditioned on lower noise. Additionally, we enable the direct training of diffusion models on symbolic music without relying on an additional VAE, by transforming discrete MIDI to pianorolls, which resemble image formats with time on the horizontal axis and pitch on the vertical axis.

Our key contributions can be summarized as follows: 
\begin{itemize}
\item To our knowledge, this is the first work to explicitly optimize the diffusion U-Net’s internal representations via an alignment loss that guides the model’s latent representations from an earlier layer with a higher noise level toward the representation from a later layer with a lower noise level.
\item We empirically investigate the effect of different layer pairings on self-representation alignment (SRA), and observe that moderately deep layers tend to provide more effective guidance to early layers.
\item The proposed diffusion model with SRA outperforms the current state-of-the-art for video-to-music generation, achieving stronger video–music alignment as measured by retrieval precision metrics.
\end{itemize}

\section{Related Work}
\label{sec:format}
\subsection{Video-to-Music Generation}
Generating music conditioned on videos is a challenging multimodal task. Early methods were typically restricted to simple scenarios such as dance or instrumental performance videos. For example, Foley Music developed by Gan et al. \cite{b6} can generate music that aligns with the body motions in the videos. Although CMT (Controllable Music Transformer) \cite{b7} extends background music generation to freestyle videos, it primarily establishes the rhythmic correspondence between the videos and the generated music, lacking abstract alignment such as emotion or style. In contrast, V-MusProd \cite{b8} captures semantic, color, and motion features from videos to guide the music generation process, thereby enabling higher-level semantic alignment between video content and background music. SONIQUE \cite{b9} employs a diffusion model for video-to-music generation, but it merely allows for the generation of audio music and struggles with precise temporal alignment. Diff-BGM \cite{b10} is the first diffusion-based framework for symbolic music generation conditioned on videos. However, it inherits a common limitation of diffusion models, suboptimal internal representation quality, which can hinder the generation performance.

\subsection{Representation Alignment}

Recently, a couple of studies have highlighted the benefits of learning high-quality internal representations in improving generation performance of diffusion models. However, standard diffusion models’ training objective does not explicitly optimize for internal latent representations. To enhance representation learning, REPresentation Alignment (REPA) \cite{b5} aligns internal states of diffusion transformers with a large-scale pre-trained representation model such as DINOv2 \cite{b11}. However, the heavy reliance on external representation models introduces additional complexity, significantly constraining its practical application. Although diffusion transformers have been shown to achieve SRA that guides representation learning without external components \cite{b12}, this does not directly apply to diffusion U-Nets. Unlike diffusion transformers, which maintain a uniform resolution across all layers, a diffusion U-Net’s downsampling operations produce representations of varying sizes. This leads to spatial dimension mismatches between layers, presenting substantial challenges for SRA in diffusion U-Nets. Our method, to the best of our knowledge, enables the first SRA in diffusion U-Nets. It can effectively generate music from videos, obtaining consistently promising evaluation results.

\section{Method}
\label{sec:pagestyle}

\begin{figure*}
\centering
\includegraphics[width=0.8\textwidth]{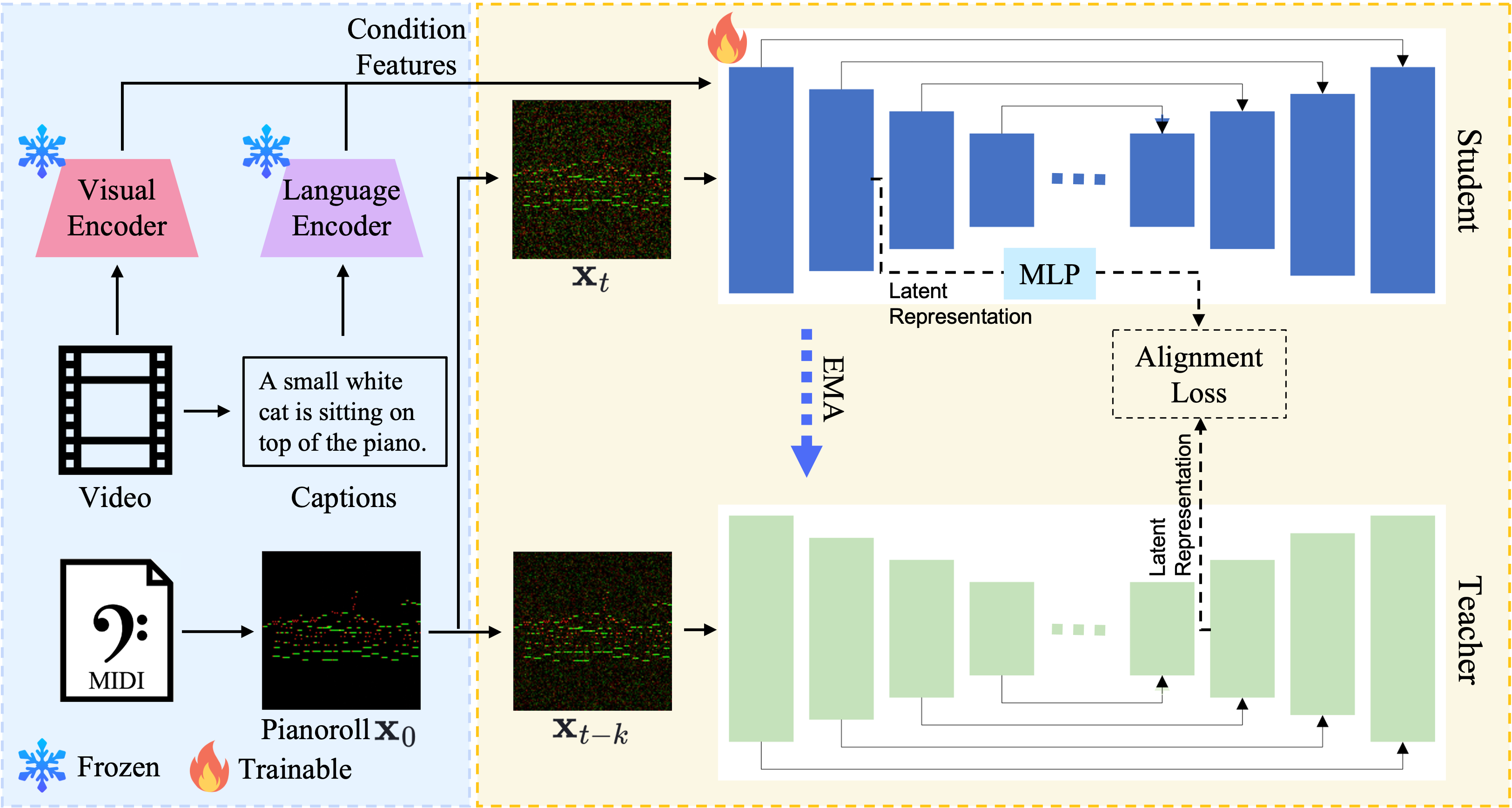}
\caption{\label{fig:MusicDiffusion}Our model aligns the student’s weaker early-layer representation conditioned on higher noise to the teacher’s later-layer latent representation conditioned on lower noise. We apply a stop-gradient operator to the teacher so gradients flow only through the student, while the teacher’s weights are updated using an exponential moving average (EMA) of the student’s parameters.}
\end{figure*}

Diffusion models \cite{b20} consist of a forward process and a reverse process. Specifically, given a clean sample $\mathbf{x}_{0}$, the forward diffusion process incrementally corrupts it by adding Gaussian noise over $T$ timesteps, producing a sequence of increasingly noisy latent variables  $\mathbf{x}_{1}$, \ldots, $\mathbf{x}_{T}$:
\begin{equation}
q\left(\mathbf{x}_{t} \mid \mathbf{x}_{t-1}\right)=\mathcal{N}\left(\mathbf{x}_{t} ; \sqrt{1-\beta_{t}} \mathbf{x}_{t-1}, \beta_{t} \mathbf{I}\right)
\end{equation}
\begin{equation}
q\left(\mathbf{x}_{1: T} \mid \mathbf{x}_{0}\right)=\prod_{t=1}^{T} q\left(\mathbf{x}_{t} \mid \mathbf{x}_{t-1}\right)
\end{equation}
where $\beta_{t}$ is a predefined variance schedule. After sufficient number of steps, $\mathbf{x}_{T}$ approaches an isotropic Gaussian noise. A neural network is then trained to approximate the reverse process, progressively denoising $\mathbf{x}_{T}$ to the original data distribution. 
\begin{equation}
p_{\theta}\left(\mathbf{x}_{t-1} \mid \mathbf{x}_{t}\right)=\mathcal{N}\left(\mathbf{x}_{t-1} ; \mu_{\theta}\left(\mathbf{x}_{t}, t\right), \mathbf{\Sigma}_{\theta}\left(\mathbf{x}_{t}, t\right)\right)
\end{equation}
\begin{equation}
p_{\theta}\left(\mathbf{x}_{0: T}\right)=p\left(\mathbf{x}_{T}\right) \prod_{t=1}^{T} p_{\theta}\left(\mathbf{x}_{t-1} \mid \mathbf{x}_{t}\right)
\end{equation}

The model is typically trained using a simple mean squared error objective:
\begin{equation*}
\mathcal{L}_{\mathrm{diff}}=\mathbb{E}_{\mathbf{x}_{0}, c, \epsilon, t}\left[\left\|\epsilon-\epsilon_{\theta}\left(\sqrt{\bar{\alpha}_{t}} \mathbf{x}_{0}+\sqrt{1-\bar{\alpha}_{t}} \epsilon, t, c\right)\right\|^{2}\right] \tag{5}
\end{equation*}
where $c$ is the condition feature, $\epsilon_{\theta}$ denotes the denoising network, $\alpha_{t}=1-\beta_{t}$, and $\bar{\alpha}_{t}=\prod_{i=1}^{t} \alpha_{i}$.

Recent work \cite{b5} has shown that learning good internal representation can enhance the performance of diffusion transformers. 
We propose an advanced diffusion U-Net with SRA to tackle the video-to-music generation task, as shown in Fig. \ref{fig:MusicDiffusion}. We introduce a teacher network that shares the same architecture as the trainable model (student) but updates its weights using an exponential moving average (EMA) of the student's weights. SRA is achieved without relying on auxiliary representation components by aligning the student's latent representation produced in its earlier layer conditioned on higher noise with that from the teacher's later layer conditioned on lower noise. We use the latent representation from the EMA model as the supervision target, since directly using the same model's output as a supervision signal may lead to training instability \cite{b13,b14}.

Let $f$ denote the trainable \emph{student} and $f_{*}$ the \emph{teacher}. 
Given a noisy $\mathbf{x}_{t}$, timestep $t$, and condition $c$, the $m$-th student layer produces the latent representation $\mathbf{y} \;=\; f^{m}(\mathbf{x}_{t},\, t,\, c)$. At a lower noise level, the teacher’s $n$-th layer yields $\mathbf{y}_{*} \;=\; f_{*}^{n}(\mathbf{x}_{t-k},\, t-k,\, c)$, where $m\le n$, $k\ge 0$ denotes the timestep interval between the student and teacher inputs, $0\le (t-k) < t_{\max}$, and $t_{\max} = 1000$. An excessively large time interval can cause the model to prioritize the alignment loss at the expense of generative aspects. In our experiments, we set the time interval to $0\text{--}200$, consistent with \cite{b12}. However, aligning the two features $\mathbf{y}$ and $\mathbf{y}_{*}$ is obstructed by a spatial resolution mismatch caused by the U-Net's downsampling. To address this, we first scale the student feature to match the teacher's spatial resolution and then pass the scaled representation $\mathbf{h}$ through a lightweight multilayer perceptron (MLP) head $j_{\psi}$ to obtain the student feature's projection $j_{\psi}(\mathbf{h})$. Eventually, we enforce SRA by aligning the projected student feature with the teacher’s output representation $\mathbf{y}_{*}$:
\begin{equation*}
\mathcal{L}_{\mathrm{ra}}=\mathbb{E}_{\mathbf{x}_{t}, t, c}\left[ \operatorname{dist}\left(\mathbf{y}_{*}, j_{\psi}\left(\mathbf{h}\right)\right)\right] \tag{6}
\end{equation*}
where $\operatorname{dist}$ denotes a distance calculation function, and we adopt smooth-$\ell_1$ objective for this function following \cite{b12}. Therefore, the total training objective includes diffusion and SRA losses:
\begin{equation*}
\mathcal{L}=\mathcal{L}_{\mathrm{diff}}+\lambda \mathcal{L}_{\mathrm{ra}} \tag{7}
\end{equation*}
where $\lambda > 0$ balances the trade-off between the generation objective and the SRA objective.

We use a pre-trained video encoder \cite{b15} to extract visual features $F_{v}$ from the videos. Additionally, a pre-trained BLIP model \cite{b16} is employed to generate the videos' captions which are then fed to a pre-trained language encoder \cite{b17} to obtain language features $F_{l}$. Subsequently, the proposed diffusion model utilises these extracted visual and language features as conditions to produce desired pianorolls that temporally and semantically match the video contents. This conditioning setup follows the baseline \cite{b10} to ensure that performance gains arise from the self-representation alignment rather than other components. The generated pianorolls can be converted back to MIDI files and eventually be rendered to audio. To ensure the alignment between video and music, we incorporate a cross-attention mechanism, where the noisy pianoroll latent representation serves as Query, and the video feature acts as Key and Value.

\section{EXPERIMENTS}
\label{sec:typestyle}
\subsection{Dataset}
We use the BGM909 \cite{b10} dataset containing 909 MIDI files of piano music aligned semantically and temporally with corresponding videos. The symbolic music pieces are sliced into 8-bar long segments. We represent each segment as an image-like pianoroll with a shape of $(2, 128, 128)$. The two channels separate onsets from sustains, making note boundaries explicit. The pianoroll representation's horizontal axis corresponds to 128 time steps (8 bars) and the vertical axis is the pitch dimension spanning from 0 to 127. Our diffusion model is trained on the pianorolls. 

\subsection{Experimental Setup}
We set the number of diffusion steps to $N = 1000$. The proposed model is trained for 100 epochs using the Adam optimiser with a fixed learning rate of $5 \times 10^{-5}$. For visual feature extraction, we adopt the official pre-trained VideoCLIP \cite{b15} model, while BLIP \cite{b16} is used for the generation of video captions which are then encoded by the official pre-trained language model \cite{b17} to extract language features. Following previous work \cite{b10}, both the visual and language encoders remain frozen during training to ensure a fair comparison.

\subsection{Results \& Discussion}

\begin{table*}
\caption{Evaluation of music quality and video–music correspondence for different models. $m$--$n$ (e.g., 3-15) denotes aligning the representation from the student's $m$-th layer (e.g., 3) with that from the $n$-th layer (e.g., 15) of the teacher. P is retrieval precision, higher values ($\uparrow$) are better. For PCHE and GPS, scores closer to ($\rightarrow$) the real data indicate better results.}
\centering
\renewcommand{\arraystretch}{1.45}
\resizebox{0.6\textwidth}{!}{
\footnotesize
\begin{tabular}{cccccc}
\specialrule{1.0pt}{0pt}{0pt}
 
\multirow{2}{*}{Model} & \multicolumn{2}{c}{Music Quality} & \multicolumn{3}{c}{Video-Music Alignment}            \\ \cline{2-6}
& PCHE$\rightarrow$ & GPS$\rightarrow$ & P@5$\uparrow$ & P@10$\uparrow$ & P@20$\uparrow$ \\ \hline
 
Real Data (BGM909)  & 2.717                & 0.708                                  & --          & --             & --             \\ 
 
Diff-BGM    & 2.756                & 0.674                                  & 12.1\%           & 21.5\%           & 39.8\%             \\ 

U-SRA (3-15) & 2.810                & 0.618                                  & 14.9\%            & 24.9\%            & 43.6\%             \\

U-SRA (5-15) & 2.715                & 0.708                                  & 14.9\%           & 25.7\%           & 43.7\%             \\

U-SRA (5-16) & 2.727       & 0.686                         & 15.0\%      & 25.2\%       & 43.0\%    \\

U-SRA (5-20) & 2.725                & 0.613                                  & 14.9\%           & 25.4\%           & 42.4\%             \\
\specialrule{1.0pt}{0pt}{0pt}
\end{tabular} 
}

\label{tab:model_comparison}
\end{table*}

\begin{figure}
\centering
\includegraphics[width=0.48\textwidth]{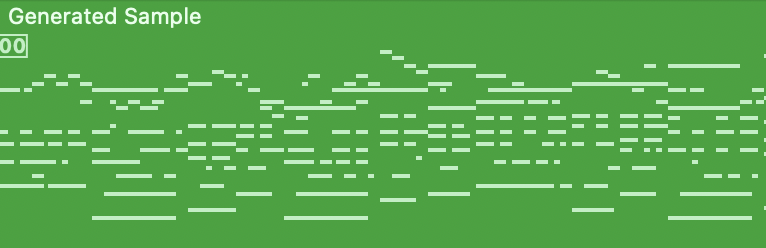}

\caption{\label{fig:music samples}A pianoroll sample generated by our U-SRA model, with pitch on the vertical axis and time on the horizontal axis.}
\end{figure}

We generated background symbolic music conditioned on the videos from the BGM909 test set. Fig. \ref{fig:music samples} shows a pianoroll segment generated by our U-SRA model. For demonstrations, please visit our project website: https://usramusic.github.io. The symbolic music generated by our model and the other video-to-music methods were evaluated using a set of metrics. Pitch Class Histogram Entropy (PCHE) \cite{b7} quantifies the music's quality in tonality by measuring the uncertainty in the distribution of pitch classes and Grooving Pattern Similarity (GPS) \cite{b7} assesses the rhythmicity of music. While these two metrics are used for evaluating music quality, we also employ a retrieval precision to assess whether the generated music matches the given video semantically and temporally. Specifically, for each generated piece of music, we construct a candidate set consisting of its corresponding ground-truth sample and $M-1$ randomly selected additional pieces. We extracted music features for each music using musicnn \cite{b18} and then computed cosine similarity between the generated track and all candidates. If the ground truth is ranked within the top-$K$ results, the generated music is considered aligned with the video. Averaging over all generated samples yields the precision score P@$K$, where we set $M=64$ and $K \in \{5, 10, 20\}$.

Non-diffusion approaches such as CMT \cite{b7} use music representations that differ fundamentally from pianorolls (e.g., event-based tokens), making it difficult to disentangle model improvements from representation effects. To enable fair and controlled comparisons, we utilize the diffusion-based Diff-BGM as our primary baseline, as it represents the current state-of-the-art for pianoroll generation conditioned on videos. Table \ref{tab:model_comparison} presents the comprehensive evaluation results for different background music generative methods, where PCHE and GPS scores closer to real data suggest higher music quality, and higher precision values are better video-music consistency. Since U-SRA (5-15) shows better performance than U-SRA (3-15), we adopt the student’s 5th layer in the subsequent experiments. Our U-SRA (5-15) model consistently outperforms Diff-BGM in both music quality metrics and video–music correspondence. These gains demonstrate that U-SRA produces higher-quality music and more effectively captures the correspondence between video and music. It is worth noting that our model obtains PCHE and GPS values which are practically the same as those of real data. The promising results verify the effectiveness of our method for tackling video-to-music generation by learning meaningful internal representations in the diffusion model via SRA.

Additionally, we conduct ablation studies to examine the effect of using different layers of student and teacher for SRA. We observe that aligning the representation from the student’s early 5th layer with that from the teacher's late 15th layer yields the best results. This is consistent with prior findings \cite{b5} that the initial few layers' weak representations require the guidance to obtain semantically meaningful features for subsequent generation. Moreover, intermediate layers are likely to provide semantically rich representations that are well-suited for guiding early student layers, while the model's excessively deep layers may shift their focus toward generating images with high-frequency details. Therefore, using the teacher’s 16th or 20th layer appears to provide less effective representation guidance than the 15th layer.

\begin{table*}
\caption{Subjective evaluation of different generative models for video background music generation, with higher values indicating better performance.}
\centering
\renewcommand{\arraystretch}{2.0}
\footnotesize
\begin{tabular}{cccccc}
\specialrule{0.8pt}{0pt}{0pt}
Model & Humanness & Richness & Rhythm & Harmony & Overall Preference \\ \hline
Diff-BGM & 60.32 $\pm$ 14.34 & 64.49 $\pm$ 14.01 & 54.26 $\pm$ 13.93 & 63.93 $\pm$ 13.99 & 56.61 $\pm$  13.22  \\
U-SRA (5-15) & 67.81 $\pm$ 11.39 & 62.35 $\pm$ 16.96 & 74.16 $\pm$ 10.45 & 72.40 $\pm$ 12.01 & 69.68 $\pm$  11.73  \\
U-SRA (5-16) & 57.96 $\pm$ 15.90 & 55.21 $\pm$ 15.11 & 65.49 $\pm$ 12.54 & 68.07 $\pm$ 14.66 & 62.82 $\pm$  11.52  \\  
\specialrule{0.8pt}{0pt}{0pt}
\end{tabular}%

\label{tab:my-table}
\end{table*}

Moreover, a subjective listening test was conducted to compare the different methods, including the U-SRA variants and Diff-BGM. Since U-SRA (5–15) and (5–16) exhibit the most competitive overall performance among the ablation variants in the objective metrics, we select them for subjective evaluation. A total of 25 participants took part in the study. Each model generated three symbolic music excerpts, resulting in nine samples in total. The symbolic outputs were rendered into audio for evaluation. All samples were presented in a randomized order, with their generating models anonymized. Participants were asked to listen to each excerpt and provide ratings on a 100-point scale across five subjective criteria \cite{b19}: Humanness, Harmony, Rhythm, Richness, and Overall Preference. Humanness measures the extent to which the generated music resembles compositions created by humans; Richness evaluates the level of diversity in the music; Rhythm examines the arrangement and pattern of sounds and silences within the music; Harmony reflects whether notes played simultaneously create a consonant and musically coherent sound. Table \ref{tab:my-table} shows the results of the subjective listening test. The U-SRA (5–15) model achieves the highest scores on most subjective metrics, while exhibiting Richness comparable to Diff-BGM. A clear consistency between objective and subjective evaluations is observed. In particular, U-SRA (5–15) attains the highest perceptual rhythm score, which corresponds to its improved GPS value. This suggests that the proposed U-SRA method not only enhances quantitative measures of rhythmicity, but also translates into perceptually meaningful improvements.



\section{Conclusion and Future Work}
\label{sec:refs}

We present a diffusion U-Net with self-representation alignment and use it to address the video-to-music generation task. By aligning the latent representation from the earlier layer conditioned on higher noise with that from the later layer conditioned on lower noise, our proposed model progressively strengthens representation learning without relying on external representation components. The extensive evaluation results suggest that the proposed model is capable of generating symbolic music with high music quality and strong alignment between the video and music, outperforming the state-of-the-art method. Although U-SRA does not rely on dataset-specific assumptions, applying it to additional video–music datasets would further validate its generalization capability, which we leave for future work.




\vspace{12pt}


\begin{thebibliography}{00}
\bibitem{b1} Y.-B. Lin, Y. Tian, L. Yang, G. Bertasius, and H. Wang, “Vmas: Video-to-music generation via semantic alignment in web music videos,” in \emph{Proc. IEEE/CVF Winter Conf. Applications of Computer Vision (WACV)}, 2025, pp. 1155--1165.
\bibitem{b2} S. Sulun, M. E. P. Davies, and P. Viana, “Symbolic music generation conditioned on continuous-valued emotions,” \emph{IEEE Access}, vol. 10, pp. 44617--44626, 2022.
\bibitem{b3} Y. Huang, A. Ghatare, Y. Liu, Z. Hu, Q. Zhang, C. S. Sastry, S. Gururani, S. Oore, and Y. Yue, “Symbolic music generation with non-differentiable rule guided diffusion,” in \emph{Proc. 41st Int. Conf. Mach. Learn. (ICML)}, vol. 235, pp. 19772--19797, 2024.
\bibitem{b4} G. Mittal, J. Engel, C. Hawthorne, and I. Simon, “Symbolic music generation with diffusion models,” in \emph{Proc. 22nd Int. Soc. Music Inf. Retrieval Conf. (ISMIR)}, 2021.
\bibitem{b5} S. Yu, S. Kwak, H. Jang, J. Jeong, J. Huang, J. Shin, and S. Xie, “Representation alignment for generation: Training diffusion transformers is easier than you think,” in \emph{Proc. 13th Int. Conf. Learn. Representations (ICLR)}, 2025.
\bibitem{b6} C. Gan, D. Huang, P. Chen, J. B. Tenenbaum, and A. Torralba, “Foley music: Learning to generate music from videos,” in \emph{Proc. Eur. Conf. Comput. Vis. (ECCV)}, pp. 758--775, 2020.
\bibitem{b7} S. Di, Z. Jiang, S. Liu, Z. Wang, L. Zhu, Z. He, H. Liu, and S. Yan, “Video background music generation with controllable music transformer,” in \emph{Proc. ACM Int. Conf. Multimedia (ACM MM)}, pp. 2037--2045, 2021.
\bibitem{b8} L. Zhuo, Z. Wang, B. Wang, Y. Liao, C. Bao, S. Peng, S. Han, A. Zhang, F. Fang, and S. Liu, “Video background music generation: Dataset, method and evaluation,” in \emph{Proc. IEEE/CVF Int. Conf. Comput. Vis. (ICCV)}, pp. 15637--15647, 2023.
\bibitem{b9} L. Zhang and M. Fuentes, “Sonique: Video background music generation using unpaired audio-visual data,” in \emph{Proc. IEEE Int. Conf. Acoust., Speech Signal Process. (ICASSP)}, pp. 1--5, 2025.
\bibitem{b10} S. Li, Y. Qin, M. Zheng, X. Jin, and Y. Liu, “Diff-bgm: A diffusion model for video background music generation,” in \emph{Proc. IEEE/CVF Conf. Comput. Vis. Pattern Recognit. (CVPR)}, pp. 27348--27357, 2024.
\bibitem{b11} M. Oquab, T. Darcet, T. Moutakanni, H. Vo, M. Szafraniec, V. Khalidov, P. Fernandez, D. Haziza, F. Massa, A. El-Nouby, \emph{et al.}, “Dinov2: Learning robust visual features without supervision,” arXiv:2304.07193, 2023.
\bibitem{b12} D. Jiang, M. Wang, L. Liuzhuozheng, L. Zhang, H. Wang, W. Wei, G. Dai, Y. Zhang, and J. Wang, “No other representation component is needed: Diffusion transformers can provide representation guidance by themselves,” arXiv:2505.02831, 2025.
\bibitem{b13} X. Chen and K. He, “Exploring simple siamese representation learning,” in \emph{Proc. IEEE/CVF Conf. Comput. Vis. Pattern Recognit. (CVPR)}, pp. 15750--15758, 2021.
\bibitem{b20} J. Ho, A. Jain, and P. Abbeel, “Denoising diffusion probabilistic models,” in \emph{Proc. Adv. Neural Inf. Process. Syst. (NeurIPS)}, vol. 33, pp. 6840--6851, 2020.
\bibitem{b14} J.-B. Grill, F. Strub, F. Altché, C. Tallec, P. Richemond, E. Buchatskaya, C. Doersch, B. A. Pires, Z. Guo, M. G. Azar, \emph{et al.}, “Bootstrap your own latent: A new approach to self-supervised learning,” in \emph{Proc. Adv. Neural Inf. Process. Syst. (NeurIPS)}, vol. 33, pp. 21271--21284, 2020.
\bibitem{b15} H. Xu, G. Ghosh, P.-Y. Huang, D. Okhonko, A. Aghajanyan, F. Metze, L. Zettlemoyer, and C. Feichtenhofer, “VideoCLIP: Contrastive pre-training for zero-shot video-text understanding,” in \emph{Proc. Conf. Empirical Methods Natural Lang. Process. (EMNLP)}, 2021.
\bibitem{b16} J. Li, D. Li, C. Xiong, and S. Hoi, “Blip: Bootstrapping language-image pre-training for unified vision-language understanding and generation,” in \emph{Proc. Int. Conf. Mach. Learn. (ICML)}, pp. 12888--12900, 2022.
\bibitem{b17} J. Devlin, M.-W. Chang, K. Lee, and K. Toutanova, “Bert: Pre-training of deep bidirectional transformers for language understanding,” in \emph{Proc. Conf. North Amer. Chapter Assoc. Comput. Linguist.: Hum. Lang. Technol. (NAACL-HLT)}, pp. 4171--4186, 2019.
\bibitem{b18} J. Pons and X. Serra, “musicnn: Pre-trained convolutional neural networks for music audio tagging,” arXiv:1909.06654, 2019.
\bibitem{b19} J. Zhang, G. Fazekas, and C. Saitis, “Mamba-diffusion model with learnable wavelet for controllable symbolic music generation,” in \emph{Proc. Int. Joint Conf. Neural Netw. (IJCNN)}, pp. 1--7, 2025.





\end{thebibliography}
\end{document}